\def\year{2022}\relax
\documentclass[letterpaper]{article} 
\usepackage{aaai22}  
\usepackage{times}  
\usepackage{helvet}  
\usepackage{courier}  
\usepackage[hyphens]{url}  
\usepackage{graphicx} 
\usepackage{natbib}  
\usepackage{caption} 
\DeclareCaptionStyle{ruled}{labelfont=normalfont,labelsep=colon,strut=off} 
\usepackage{algorithm}
\usepackage{algorithmic}

\usepackage{amsmath}
\usepackage{amssymb}
\usepackage{subcaption}
\usepackage{verbatim}
\usepackage{multirow}
\usepackage{booktabs}
\usepackage[T1]{fontenc}
\usepackage{textcomp}
\usepackage{siunitx}
\usepackage[pagebackref=true,breaklinks=true,colorlinks,bookmarks=false]{hyperref}

\usepackage{pdfpages}

\newcommand{\eg}{\textit{e.g., }}
\newcommand{\ie}{\textit{i.e., }}
\newcommand{\argmin}{\arg\!\min} 
\usepackage[utf8x]{inputenc}
\usepackage{newfloat}
\usepackage{listings}
\floatstyle{ruled}
\newfloat{listing}{tb}{lst}{}
\floatname{listing}{Listing}
\title{Improving 360$^\circ$ Monocular Depth Estimation \\
via Non-local Dense Prediction Transformer  \\
and Joint Supervised and Self-supervised Learning}
\author{
    Ilwi Yun\textsuperscript{\rm 1} ,
    Hyuk-Jae Lee \textsuperscript{\rm 1},
    Chae Eun Rhee \textsuperscript{\rm 2}
}
\affiliations{
 \textsuperscript{\rm 1} Seoul National University, Korea\\
 \textsuperscript{\rm 2} Inha University, Korea\\
  yuniw@capp.snu.ac.kr, hjlee@capp.snu.ac.kr, chae.rhee@inha.ac.kr
}

\begin{document}
\maketitle
\begin{abstract}
Due to difficulties in acquiring ground truth depth of equirectangular (360$^\circ$) images, the quality and quantity of equirectangular depth data today is insufficient to represent the various scenes in the world. Therefore, 360$^\circ$ depth estimation studies, which relied solely on supervised learning, are destined to produce unsatisfactory results. Although self-supervised learning methods focusing on equirectangular images (EIs) are introduced, they often have incorrect or non-unique solutions, causing unstable performance. 
In this paper, we propose 360$^\circ$ monocular depth estimation methods which improve on the areas that limited previous studies. First, we introduce a self-supervised 360$^\circ$ depth learning method that only utilizes gravity-aligned videos, which has the potential to eliminate the needs for depth data during the training procedure. 
 Second, we propose a joint learning scheme realized by combining supervised and self-supervised learning. The weakness of each learning is compensated, thus leading to more accurate depth estimation. Third, we propose a non-local fusion block, which can further retain the global information encoded by vision transformer when reconstructing the depths. With the proposed methods, we successfully apply the transformer to 360$^\circ$ depth estimations, to the best of our knowledge, which has not been tried before. On several benchmarks, our approach achieves significant improvements over previous works and establishes a state of the art.

\end{abstract}

\section{Introduction}
\label{Introduction}
Recently, research interest in processing equirectangular (360$^\circ$) images has increased as virtual reality enters the limelight. 
Equirectangular images (EIs) have advantages over traditional rectilinear images (RIs) in that they enable a \ang{360} field of view. This benefit, however, complicates the acquisition of ground truth depths. Aside from the technical difficulties associated with \ang{360} depth scanners, one practical difficulty is that sensors would be visible from the \ang{360} RGB cameras, leading to partially obscured images \cite{Lowcost, omnidepth}. Moreover, to acquire diverse and realistic synthesized data, numerous things should be set exquisitely, which often requires professional designers and tools \cite{structure3d}. Due to such problems, the quality and quantity of equirectangular depth data today is insufficient to represent fully the various scenes in the world.                                                                                          
 Therefore, learning 360$^\circ$ depths in a supervised manner is destined to produce unsatisfactory results because the performance of supervised learning is highly dependent on the dataset.
To overcome the lack of data, learning 360$^\circ$ depths in a self-supervised manner has been attempted. However, previous methods require either calibrated stereo EI pairs \cite{EBS,svsyn,360sdnet} or conversion to cubemap projection \cite{360self}, both of which have limitations with regard to further improvements. Moreover, self-supervised learning often delivers incorrect or non-unique solutions (\eg light reflected object), which cause unstable performance.
 
\begin{figure}[t]
\centering
\begin{minipage}{.42\textwidth}
\begin{subfigure}{\linewidth}
\includegraphics[width=\linewidth]{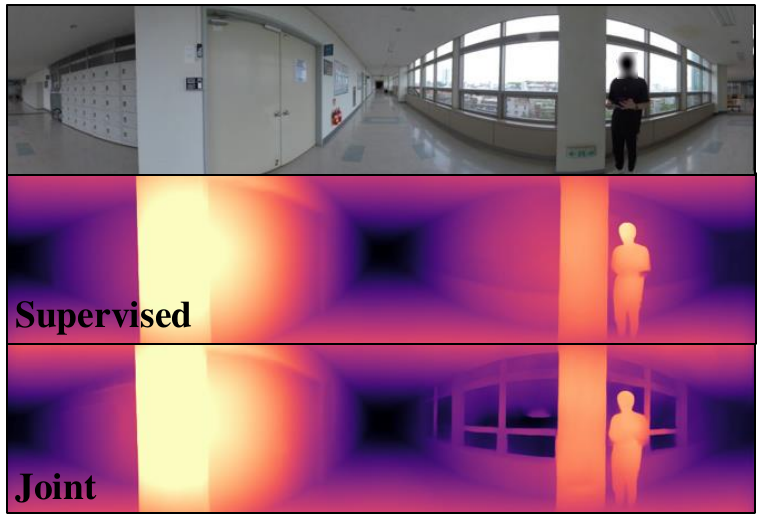}
\end{subfigure}
\end{minipage}%
\caption{The effect of joint learning. Unlike the network trained via supervised learning only (middle row), the network trained via joint learning is able to distinguish windows from the walls (bottom row)}
\label{fig:intro}
\end{figure}
 
In this paper, we propose 360$^\circ$ monocular depth estimation methods which improve on the areas that limited previous studies. First, we propose a self-supervised method for the learning of depth that only utilizes gravity-aligned video sequences, which has the potential to eliminate the needs of depth data during the training procedure. Similar to prior work \cite{svsyn}, we utilize the relationships between consecutive scenes but improve it through consistency between depths.
 Second, we propose a joint learning scheme realized by combining supervised and self-supervised learning. 
  Despite the limitations of each learning scheme, all previous works on the 360$^\circ$ depth estimation, to the best of our knowledge, have relied solely on either supervised or self-supervised learning. 
 We show that the joint learning improves the unstable performance of self-supervised learning as well as the incorrect prediction of supervised learning caused by data scarcity, as visualized in Figure \ref{fig:intro}. Third, we propose a non-local fusion block which improves on the areas missed by vision transformers for dense prediction. Through non-local operations, global information encoded by a transformer can be further retained when reconstructing the depths. Under a challenging environment for a transformer (\ie lack of a large-scale dataset), we were able to train the vision transformer successfully using the features learned from depth of RIs. To the best of our knowledge, this is the first work applying transformers successfully to 360$^\circ$ depth estimation. Our approaches achieve significant improvements over previous works on several benchmarks, thus establishing a state of the art.



\section{Background and Related work}
\subsection{EI geometry}
\begin{figure}[h]
\centering

\begin{minipage}{.18\textwidth}
\begin{subfigure}{\linewidth}
\includegraphics[width=\linewidth]{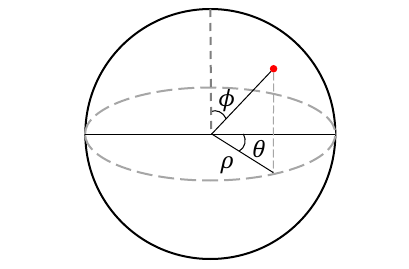}
\caption{Sphere}
\end{subfigure}
\end{minipage}%
\begin{minipage}{.18\textwidth}
\begin{subfigure}{\linewidth}
\includegraphics[width=\linewidth]{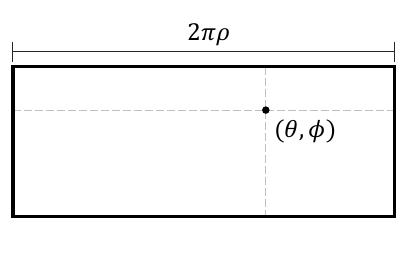}
\caption{Equirectangular}
\end{subfigure}
\end{minipage}%
\caption{Equirectangular geometry}
\label{fig:ERP_geometry}
\end{figure}

Although EIs appear to be two dimensional (2D) images, EIs and RIs are different in many ways. EIs are generated by flattening the rays projected on a three dimensional (3D) sphere, whereas RIs are generated by directly projecting rays on a 2D plane. 
 Therefore, EIs are 3D images despite their 2D structure. The spherical coordinates $(\theta,\phi,\rho)$ are often used instead of the pixel coordinates $(x,y)$ for this reason, and relationship between them is illustrated in Figure \ref{fig:ERP_geometry}.
 Each value of $\theta \in (\ang{0},\ang{360})$ and $\phi \in (\ang{0},\ang{180})$ represents the latitude and longitude of an EI, and $\rho$ denotes the radius of the sphere.  
 Further, the spherical coordinates can be converted to Cartesian coordinates $(X_c,Y_c,Z_c)$ by Eq.\ref{sph2cart}

 \begin{equation}
\label{sph2cart}
\begin{cases} 
X_c = \rho \cdot \sin(\phi) \cdot \cos(\theta) \\
Y_c = \rho \cdot \sin(\phi) \cdot \sin(\theta) \\
Z_c = \rho \cdot \cos(\phi)
\end{cases}
\end{equation}

Meanwhile, rotations on EIs are defined as the yaw, pitch and roll. Due to undesirable visual changes \cite{hohonet, 360align}, a gravity-aligned structure (\ie with the roll and pitch set to 0$^\circ$) is generally assumed in equirectangular depth benchmarks \cite{stanford,matterport,omnidepth,structure3d} and in recent studies \cite{Slicenet, hohonet}. If captured images/videos are not gravity-aligned, they can be calibrated afterwards \citep{uprightnet,360align}.    
 
\subsection{Supervised 360$^\circ$ depth estimation}
 Omnidepth \cite{omnidepth} presents a 3D60 dataset (Matterport3D, Stanford3D and SunCG) by re-rendering previous 360$^\circ$ data [\eg Matterport \cite{matterport}, Stanford \cite{stanford}], which are now commonly used for training 360$^\circ$ depths. Bifuse \cite{bifuse} jointly uses cubemap projected images with EIs to improve the performance. SliceNet \cite{Slicenet} splits the inputs and recovers them through long short-term memory \cite{LSTM} to retain the global information. HoHoNet \cite{hohonet} improves the performance and computational efficiency by focusing on the per-column information of the gravity-aligned EIs.
 Recently, multi-task learning among the depth, layout and semantics was attempted to improve performance outcomes.
 \cite{geo_depth} regularizes the depth considering the layout, while \cite{jointdepth} train the layout, semantic and depth simultaneously. 

\subsection{Self-supervised 360$^\circ$ depth estimation}
\label{sec:related_work_self}
\begin{figure}[h]
\centering
\begin{minipage}{.15\textwidth}
\begin{subfigure}{\linewidth}
\includegraphics[width=\linewidth]{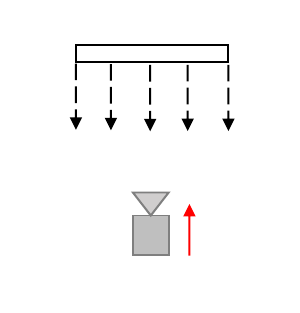}
\caption{Rectilinear}
\end{subfigure}
\end{minipage}%
\begin{minipage}{.15\textwidth}
\begin{subfigure}{\linewidth}
\includegraphics[width=\linewidth]{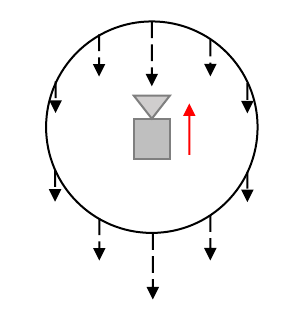}
\caption{Equirectangular}
\end{subfigure}
\end{minipage}%
\caption{Difference in movements of a scene}
\label{fig:rendering}
\end{figure}

 Self-supervised depth learning has been widely attempted for RIs based on the following intuition: The closer the object is to the camera, the greater the change in the object's position when the camera moves \cite{unsuperdepth,monodepth,monodepth2,wilddepth}. 
However, that intuition is not applied to EIs due to the different geometry, as shown in Figure \ref{fig:rendering}.  
When the camera moves forward (denoted by the red arrows), the relative movements of objects in the scenes of RIs are represented by the dotted arrows in Figure \ref{fig:rendering} (a). The movement of objects only depends on the camera movements (direction) and corresponding depths (magnitude). 
The movement of objects in EIs, however, is also affected by the positions of the objects in EIs. As the camera moves forward, the objects in front of the camera become closer, while those of opposite side become further away, as shown in Figure \ref{fig:rendering} (b). Because more variables control the objects of EIs, learning 360$^\circ$ depths using a self-supervision becomes more difficult.

In addition, self-supervised depth learning often has incorrect or non-unique solutions in some cases. Light reflected objects, which are not predictable using only the depth and camera motion, is one such example. 
These objects cause the network to output wrong depth values because the light reflection is controlled by light sources, not depths (Refer to Technical Appendix for more examples). Although several attempts have been made to remove those kinds of intractable objects during the training process \cite{monodepth2,wilddepth,mirror3d}, there remain objects controlled by numerous variables and, not merely by depth and camera motions, which makes self-supervised learning a challenge.

  \begin{figure*}[!t]
\centering
\begin{minipage}{.9\textwidth}
\begin{subfigure}{\linewidth}
\includegraphics[width=\linewidth]{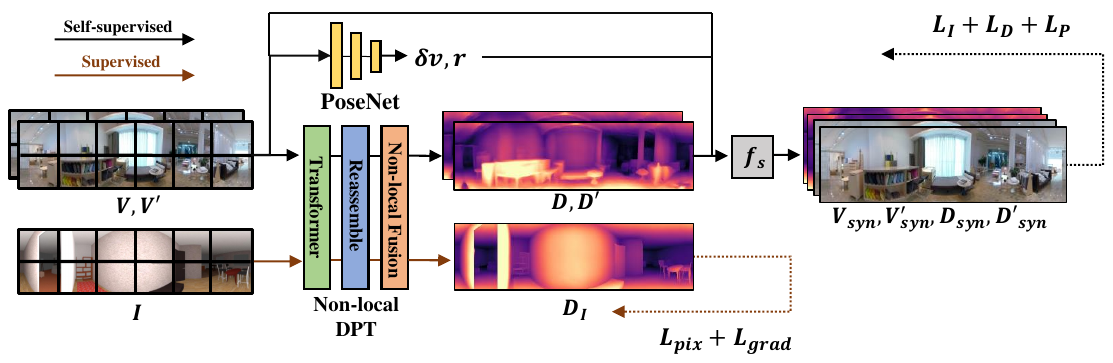}
\end{subfigure}
\end{minipage}%
\caption{Overall architecture}
\label{fig:Overall_arch}
\end{figure*}

360SD-Net \cite{360sdnet} and SvSyn \cite{svsyn} use stereo EI pairs as input for EI depth training data to simplify the relationship between the depth, image and camera motions.  However, 360$^\circ$ field of view makes it difficult to acquire stereo EIs using two 360$^\circ$ cameras given that each camera is captured by others, which limits the use of this method. 
EBS \cite{EBS} proposes the method that uses relatively abundant RI stereo pairs. They distort RIs considering the EI geometry, and use them as training data. However, distorted data has a restricted field of view ($<$ \ang{90}), which cannot replace the EIs fundamentally. 
 \cite{360self} transforms EIs into cubemap-projected images to alleviate the difference in the geometry. However, cube map projection not only leads to discontinuity between each of the cubemap faces which results in large errors, but also requires additional computations \cite{cubepadding,bifuse}.

\subsection{Vision transformer}
Recently, the vision transformer (ViT) network architecture was proposed for image classification \cite{vision_transformer}. In this architecture, the transformer \cite{attention}, which has been widely used in natural language processing, is adopted instead of convolution block. Images are split into multiple flattened patches and are encoded by a transformer. ViT achieves results that are comparable to or even better than those of a convolutional neural network (CNN) on image classification tasks. Furthermore, it was recently demonstrated that ViT yields notable performance improvements on various vision tasks. The dense prediction transformer (DPT) successfully applies ViT to segmentation and depth estimation tasks by upsampling the encoded features via convolution-based reassemble and fusion blocks (FB) \cite{DPT}. Reassemble blocks reassemble the encoded features into 3D features $R^s$, whereas fusion blocks upsample the $R^s$ into fused features $F^s$.
Unlike a CNN, however, the transformer lacks inductive bias, necessitating  large-scale dataset. Under an environment with an insufficient dataset, the performance of the transformer becomes worse than that of a CNN \cite{vision_transformer}. Therefore, several attempts have been made to alleviate data dependencies through multiple known techniques, such as knowledge distillation \cite{deit}.

\section{Proposed method}
\subsection{Overall architecture}
The overall structure of the proposed training procedure, as illustrated in Figure \ref{fig:Overall_arch}, is composed of two flows: self-supervised learning (black arrows) and supervised learning (brown arrows). For self-supervised learning flow, non-local DPT estimates the depth ($D, D'$) for each consecutive video scene ($V, V'$), while PoseNet predicts the camera motion ($\delta v,\delta r$) between them. Then, each scene and depth at different viewpoints are reconstructed through sampling function ($f_s$). Image consistency ($\mathcal{L}_I$), depth consistency ($\mathcal{L}_D$) and pose consistency ($\mathcal{L}_P$) losses are imposed by comparing reconstructed samples with corresponding video scenes and the estimated depths. For supervised learning flow, estimated depth ($D_I$) for image input ($I$) is compared with the ground truth ($D^g_I$). Traditional pixel-wise loss ($\mathcal{L}_{pix}$) and gradient loss ($\mathcal{L}_{grad}$) are imposed. Overall, the objective function is constructed via Eq.\ref{total}. Here, $\lambda_I$ and $\lambda_D$ are hyper-parameters balancing the supervised and self-supervised losses. Each element of Eq.\ref{total} will be described in the following sections.
   
\begin{equation}
\label{total}
\mathcal{L}_{total} = \lambda_I \cdot \mathcal{L}_I + \lambda_D \cdot \mathcal{L}_D + \mathcal{L}_P + \mathcal{L}_{pix} + \mathcal{L}_{grad}
\end{equation}

\subsection{Self-supervised losses}
\label{sec:proposed_self}
In this subsection, we introduce the self-supervised learning flow illustrated in Figure \ref{fig:Overall_arch}. 
 First, we formulate the relationships among the depth, camera motion and gravity-aligned 360$^\circ$ video sequences. Then, the image consistency, depth consistency and pose consistency losses are explained. \\
\par{\noindent\textbf{Relationships between consecutive scenes in video}}
 Consecutive video scenes $V \in \mathbb{R}^{3 \times H \times W }$ and $ V' \in \mathbb{R}^{3 \times H \times W}$ can be expressed as spherical coordinate ($\theta, \phi, \rho$) and ($\theta', \phi',\rho'$), respectively. Here, each of the $\rho$ and $\rho'$ values can be considered as estimated depths, which are denoted correspondingly as $D \in \mathbb{R}^{1 \times H \times W }$ and $D' \in \mathbb{R}^{1 \times H \times W }$. When the video proceeds from $V$ to $V'$  (\ie $V \rightarrow V'$), the translation and rotation of the camera between scenes are defined as $\delta v \in \mathbb{R}^{3 \times 1 \times 1 }$ and $\delta r \in \mathbb{R}^{3 \times 1 \times 1 }$. Because we assume that the videos are gravity-aligned, $\delta r_y \in \mathbb{R}^{1 \times 1 \times 1 }, \delta r_z \in \mathbb{R}^{1 \times 1 \times 1 } $ can be set as a constant (\ie 0). Therefore, the camera motion is simplified to four variables: $\delta v_x, \delta v_y, \delta v_z$ and  $\delta r_x$. Under this environment, Eq.\ref{consistency} is formulated, which denotes the movements of the 3D scene point between $V$ and $V'$ in Cartesian coordinates according to the camera motion and depth .
\begin{equation}
\label{consistency}
\begin{cases}
\rho \cdot \cos(\theta - \delta r_x) \cdot \sin(\phi) - \delta v_x = \rho' \cdot \cos(\theta') \cdot \sin(\phi') \\
\rho \cdot \sin(\phi) \cdot \sin(\theta - \delta r_x ) - \delta v_y = \rho' \cdot \sin(\phi') \cdot \sin(\theta') \\
\rho \cdot \cos(\phi) - \delta v_z= \rho' \cdot \cos(\phi') 

\end{cases}
\end{equation}

 By solving Eq.\ref{consistency}, a closed-form expression of Eq.\ref{Equation_solv} is obtained, representing the relationship between $V$ ($\theta, \phi, \rho$) and $V'$ ($\theta', \phi',\rho'$) for the depth and camera motion. Eq.\ref{Equation_solv} can be expressed simply using $f_s$ in Eq.\ref{simple}. 

 \begin{equation}
\label{Equation_solv}
\begin{cases}
\theta' = \tan^{-1}(\cfrac {\rho \cdot \sin(\theta- \delta r_x) \cdot \sin(\phi)- \delta v_y}{\rho \cdot \cos(\theta - \delta r_x) \cdot \sin(\phi) - \delta v_x}) \\
\phi' = \tan^{-1}(\cfrac{\rho \cdot \sin(\theta - \delta r_x) \cdot \sin(\phi) - \delta v_y}{ \sin(\theta') \cdot (\rho \cdot \cos(\phi) - \delta v_z)}) \\
\rho' = \cfrac{\cos(\phi) \cdot \rho - \delta v_z}{\cos(\phi')} 

\end{cases}
\end{equation}

\begin{equation}
\label{simple}
V'= f_s(V,D, \delta v, \delta r)
\end{equation}

\par{\noindent\textbf{Image consistency loss}}
For consecutive scenes $V \rightarrow V'$, scenes at a different view point $V'_{syn}$ can be synthesized from $V$ with Eq.\ref{simple_loss} assuming the depth and camera motions are well estimated. In the same vein, $V_{syn}$ can also be synthesized considering the reversely ordered sequences $V' \rightarrow V$. If the depth and camera motions are appropriately estimated, each synthesized frame should be identical to each corresponding scene in the video (\ie $V' = V'_{syn}$ and $V = V_{syn}$). Therefore, by regularizing the networks to synthesize images equal to scenes in the video, the networ is indirectly trained to predict proper depth and camera motions. Therefore, the image consistency loss is constructed as Eq.\ref{image consistency}, similar to the previous works \cite{svsyn}. Here, $SM$ indicates structural similarity \cite{SSIM}, $\alpha$ represents the weight parameters, $n$ indicates the number of pixels in an image. 

\begin{align}
\label{simple_loss}
\begin{gathered}
 V'_{syn}= f_s(V,D, \delta v, \delta r) \\
 V_{syn}= f_s(V',D', \text{-}\delta v, \text{-}\delta r)
 \end{gathered}
\end{align}

\begin{align}
\label{image consistency}
\begin{split}
&L_I = {1 \over n}\cdot \sum_{k=1}^{n}{[\alpha \cdot (|V' - V'_{syn}| + |V - V_{syn}|)} \\  
&+ (1 - \alpha) \cdot (|1 - SM(V,V_{syn})| + |1 - SM(V',V'_{syn})|)]
\end{split}
\end{align}

\par{\noindent\textbf{Depth consistency loss}}
Previous self-supervised learning studies focusing on EIs only considered image consistency \cite{svsyn,360sdnet}. We argue that depth consistency can also be used for regularization, which can further strengthen the training. 
Here, we introduce the depth consistency loss for EIs, inspired by \cite{monodepth}. By regarding the depth as an image, the depths of different viewpoints $D'_{syn}$ and $D_{syn}$ can be synthesized using Eq.\ref{depth syn}. If the depth and camera motion are accurately estimated, $D'_{syn}$ and $D_{syn}$ become equal to $D'$ and $D$, respectively. Therefore, similar to the image consistency loss, $\mathcal{L}_D$ can be constructed with Eq.\ref{depth consistency}. The estimated depths should be consistent across the scenes to minimize the loss, which causes the network to check the images in more detail.

\begin{equation}
\label{depth syn}
\begin{gathered}
D'_{syn} = f_s(D,D,\delta v, \delta r) \\
D_{syn} = f_s(D',D',\text{-}\delta v, \text{-}\delta r)
\end{gathered}
\end{equation}
\begin{equation}
\label{depth consistency}
L_D = {1 \over n}\cdot \sum_{k=1}^{n}{(|D' - D'_{syn}| + |D - D_{syn}|)}
\end{equation}

\par{\noindent\textbf{Pose consistency loss}}
If PoseNet $P$ is properly trained, estimated camera motions for scenes in reverse order (\ie $V \rightarrow V'$ and $V' \rightarrow V$) should also have the opposite direction. Therefore, the pose consistency loss $L_P$ can be established as Eq.\ref{pose consistency}.

\begin{equation}
\label{pose consistency}
L_P = {1 \over n}\cdot \sum_{k=1}^{4}{|P(V,V') - P(V',V)|}
\end{equation}

\subsection{Supervised losses for joint learning}
\label{sec:supervised_loss}
Although self-supervised learning on videos has the potential to offer more accurate depth estimation, it has drawbacks that should be resolved (\ie non-optimal solutions).
Supervised learning can alleviate those drawbacks because it has supervisions in the areas (\eg light reflection) which cause problems in self-supervised learning. 
Conversely, self-supervised learning can diversify the features learned from supervised learning through various video sequences, which makes the network perform well at the unexposed data. From this observation, we propose to use the supervised and self-supervised losses jointly. However, the scale of the depth value differs according to the depth-acquisition method used or the environment \cite{silog_loss,ordinal_loss,NMG_loss,midas}. Therefore, scale and shift ambiguities regarding the depth must be resolved in advance to use both losses together. If this is not done, they conflict harshly and produce even worse performances. For this reason, we initially align the scale and shift of the depth via Eq.\ref{align} utilizing schemes proposed by \cite{midas}. Considering the per-column characteristics of the gravity-aligned EIs \cite{hohonet}, the depth is aligned in a column-wise manner. Here, $D^g_I \in \mathbb{R}^{1 \times H \times W }$ indicates the ground truth depth,  $D^A_I \in \mathbb{R}^{1 \times H \times W }$ is the aligned depth, where $s,t \in \mathbb{R}^{1 \times 1 \times W }$ represents the per-column scale and shift parameters.

\begin{equation}
\begin{gathered}
\label{align}
s,t = \argmin_{s,t}(s \cdot D_I + t - D^g_I) \\
D^{A}_{I} = s\cdot D_I + t
\end{gathered}
\end{equation}

Then, we apply traditional pixel-wise loss expressed as Eq.\ref{pixel wise} and the gradient loss \cite{megadepth} of Eq.\ref{gradient}, while the gradient loss is calculated on four different scales. The gradient loss induces the network to place emphasis on the edges of the estimated depths.

\begin{equation}
\label{pixel wise}
\mathcal{L}_{pix} = {1 \over n}\cdot \sum_{k=1}^{n}{|(D^A_I - D^g_I)|}
\end{equation}

\begin{equation}
\label{gradient}
\mathcal{L}_{grad} = {1 \over n}\cdot \sum_{k=1}^{n}{|\nabla_x (D^A_I - D^g_I) + \nabla_y (D^A_I - D^g_I)|}
\end{equation}

\subsection{Non-local dense prediction transformer}
\label{sec:NLDPT}
\par
\begin{figure}[h]
\centering
\begin{minipage}{.31\textwidth}
\begin{subfigure}{\linewidth}
\includegraphics[width=\linewidth]{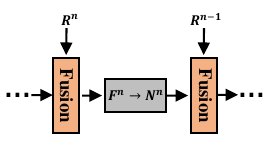}
\end{subfigure}
\end{minipage}%
\caption{Non-local fusion block}
\label{fig:NLFB}
\end{figure}

{\noindent\textbf{Non-local fusion block}} ViT has advantages over a CNN in that it can see the input images globally. For dense predictions, however, these advantages might be weakened. When upsampling encoded features via convolution-based FB \cite{DPT}, the receptive field is bounded to the convolution kernel size. Because EI contains a geometric structure (\eg a wall) which should be seen in a global manner \cite{Slicenet}, losing the global outlook may result in non-accurate depth estimations. For these reasons, we propose the use of a non-local fusion block (NLFB), which performs non-local operations \cite{non_local} on each feature going into the FB, as shown in Figure \ref{fig:NLFB}. Here, we define $F^s \in \mathbb{R}^{C \times H_s \times W_s }$ as the fused features at scale $s$ and $N^s \in \mathbb{R}^{C \times H_s \times W_s }$ as the non-local fused features. For the fused feature at the $i$th index $F^s_i$, the non-local fused feature $N^s_i$ is calculated with Eq.\ref{non_local}, where $W^s_{\theta,\phi,g} \in \mathbb{R}^{C/2 \times 1 \times 1 }$ and $W^s_{z} \in \mathbb{R}^{C \times 1 \times 1 }$ are the weight matrix to be learned and $j$ denotes the array of all possible indexes. The features of each index $i$ are reconstructed by seeing all other indexes $j$, which makes the network to continue seeing the features from a global perspective. Therefore, non-local DPT, which uses a NLFB instead of a FB, yields more accurate dense predictions.

\begin{equation}
\label{non_local}
\begin{gathered}
f(F^s_i,F^s_j) = e^{(W^s_{\theta}F_i)^T W^s_{\phi}F_j}\\
C(F^s) = \sum_{\forall j}{f(F^s_i,F^s_j)} \\
N^s_i = F^s_i + {W^s_z \over C(F^s)} \sum_{\forall j}{f(F^s_i,F^s_j)W^s_gF^s_j}
\end{gathered}
\end{equation}

\par{\noindent\textbf{Fine-tuning on EI depth}} Due to lack of a inductive bias (\eg locality assumption), a large-scale dataset is needed to train a well-performing transformer \cite{vision_transformer}. Considering the small amount of EI depth data, this disadvantage is especially critical in EI depth estimation studies. To overcome this hardship, we utilize a pre-trained model which is learned from depth of RIs based on the following observation: Depth estimations on RIs and EIs work similarly from the relative point of view. Here, we assume that two objects $A$ and $B$ are captured by RI and EI cameras, respectively. If it is perceived that $A$ is closer than $B$ in RIs, it also should be the case in EIs. This implies that features learned from depth of RIs are also useful for EI depths. Therefore, we initialize the weights of our network using the transformer trained with RI depths and fine-tune the whole network using EI depths with some additional settings. 

Although estimated depths are aligned via Eq.\ref{align} during the training procedure, we observe that scale mismatches in depth between two learning flows dominate the total loss. Before aligning depths via Eq.\ref{align}, we thus robustly adjust the scales of ground truth and the estimated depth of supervised learning flows based on the scales of the depth learned from self-supervised losses. Further, we also pre-train the pose network to minimize the negative effects of incorrect pose estimations in the early phase of training. In this way, we successfully transfer the various features learned from large-scale RI depths to the equirectangular geometry. More details are described in Technical Appendix.

\begin{table*}[t]
\renewcommand{\tabcolsep}{2mm}
\small
\centering
\caption{Quantitative comparison on 3D60 dataset using the pre-trained baselines provided by each author. Numbers in \textbf{bold} indicate the best results.}
\begin{tabular}{cc||cccc|ccc}
\hline
Dataset & Method & AbsRel & Sq rel & RMS & RMSlog & $\delta \scalebox{0.9}{ < 1.25}$ & $\delta\scalebox{0.9}{ < 1.25}^2 $ & $\delta\scalebox{0.9}{ < 1.25}^3$ \\ \hline

\multirow{5}{*}{Stanford3D} & Omnidepth   & 0.1009 & 0.0522&0.3835 &0.1434 &0.9114 &0.9855 &  0.9958 \\
& SvSyn   & 0.1003 & 0.0492 & 0.3614 & 0.1478 & 0.9096 & 0.9822 &  0.9949 \\
& Bifuse   & 0.1214 & 0.1019 & 0.5396 & 0.1862 & 0.8568 & 0.9599 &  0.9880 \\
& HoHoNet  & 0.0901 & 0.0593 & 0.4132 & 0.1511 & 0.9047 & 0.9762 & 0.9933 \\
& Ours w/ FB & 0.0669 & 0.0249 & 0.2805 & 0.1012 & 0.9652 & 0.9944 & 0.9983 \\
& Ours w/ NLFB & \textbf{0.0649} & \textbf{0.0240} & \textbf{0.2776} & \textbf{0.993} & \textbf{0.9665} & \textbf{0.9948} & \textbf{0.9983} \\ \hline \hline

\multirow{5}{*}{Matterport3D} & Omnidepth & 0.1136 & 0.0671 & 0.4438 & 0.1591 & 0.8795 & 0.9795 & 0.9950 \\
& SvSyn & 0.1063 & 0.0599 & 0.4062 & 0.1569 & 0.8984 & 0.9773 & 0.9934 \\
& Bifuse & 0.1330 & 0.1359 & 0.6277 & 0.2079 & 0.8381 & 0.9444 & 0.9815 \\
& HoHoNet & \textbf{0.0671} & 0.0417 & 0.3416 & 0.1270 & 0.9415 & 0.9838 & 0.9942 \\
& Ours w/ FB & 0.0729 & 0.0302 & 0.3089 & 0.1079 & 0.9574 & 0.9935 & 0.9980 \\
& Ours w/ NLFB & 0.0700 & \textbf{0.0287} & \textbf{0.3032} & \textbf{0.1051} & \textbf{0.9599} & \textbf{0.9938} & \textbf{0.9982} \\ \hline \hline

\multirow{5}{*}{SunCG} & Omnidepth & 0.1450 & 0.1052 & 0.5684 & 0.1884 & 0.8105 & 0.9761 & 0.9941 \\
& SvSyn & 0.1867 & 0.1715 & 0.6935 & 0.2380 & 0.7222 & 0.9427 &  0.9840 \\
& Bifuse & 0.2203 & 0.2693 & 0.8869 & 0.2864 & 0.6719 & 0.8846 & 0.9660 \\
& HoHoNet & 0.0827 & 0.0633 & 0.3863 & 0.1508 & 0.9266 & 0.9765 & 0.9908 \\
& Ours w/ FB & 0.0740 & 0.0338 & 0.3475 & 0.1073 & 0.9584 & 0.9949 & 0.9986 \\
& Ours w/ NLFB & \textbf{0.0715} & \textbf{0.0321} & \textbf{0.3401} & \textbf{0.1042} & \textbf{0.9625} & \textbf{0.9950} & \textbf{0.9986} \\ 

\hline

\end{tabular}
\label{tab:depth_results}       
\end{table*}
\begin{figure*}[!t]
\centering
\begin{minipage}{.333\textwidth}
\begin{subfigure}{\linewidth}
\includegraphics[width=.98\linewidth]{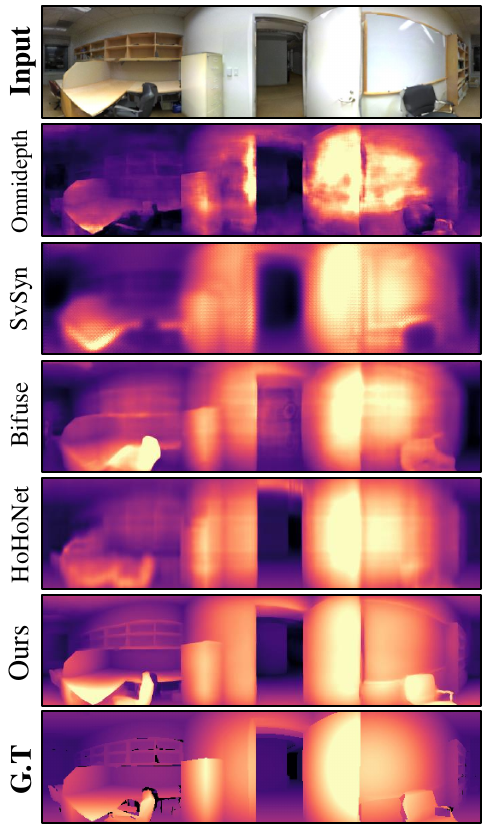}
\caption{Stanford3D}
\end{subfigure}
\end{minipage}%
\begin{minipage}{.333\textwidth}
\begin{subfigure}{\linewidth}
\includegraphics[width=.98\linewidth]{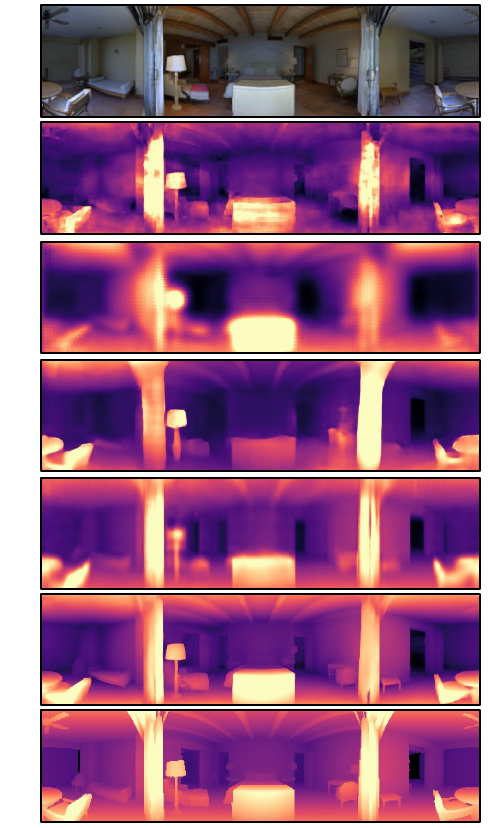}
\caption{Matterport3D}
\end{subfigure}
\end{minipage}%
\begin{minipage}{.333\textwidth}
\begin{subfigure}{\linewidth}
\includegraphics[width=.98\linewidth]{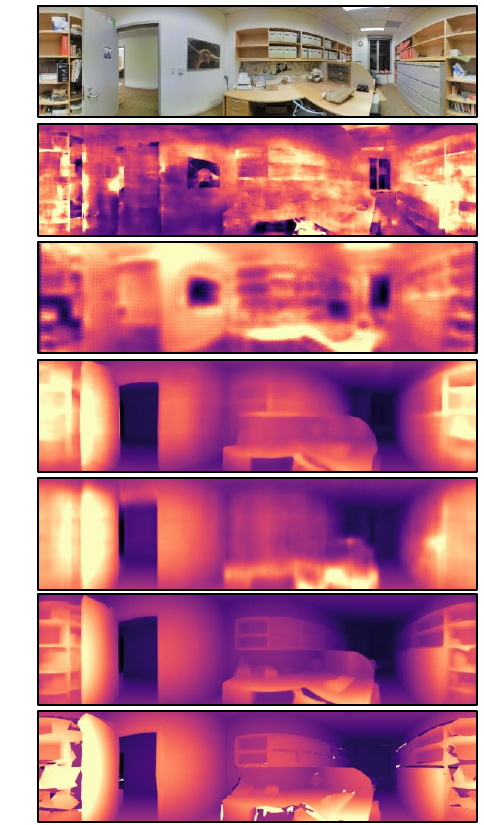}
\caption{Stanford}
\end{subfigure}
\end{minipage}%
\caption{Qualitative comparison on Stanford3D, Matterport3D and Stanford dataset. Here, Ours indicates the model using NLFB module. Additional results are included in Technical and Multimedia Appendix.}
\label{fig:depth_results}
\end{figure*} 
\section{Experiments}
\label{sec:experiment}
Experiments are composed of four parts. First, we briefly explain the experiment environment in Section \ref{Environment}. Then, we evaluate our methods under the various settings in Sections \ref{Comp_Pre2} and \ref{Comp_SOTA}. Finally, the effect of each proposed schemes is analyzed through an ablation study in Section \ref{Ablation study}. 
\subsection{Experimental setup}
\label{Environment}
\par{\noindent\textbf{Evaluation details}} For fair and reproducible experiments, we compare our results with studies that provide a pretrained model in the open-source community. 
Omnidepth \cite{omnidepth}, SvSyn \cite{svsyn}, Bifuse \cite{bifuse} and HoHoNet \cite{hohonet}, which are proven useful in numerous works, are compared. 
Stanford \cite{stanford} and 3D60 (Stanford3D, Matterport3D, SunCG) \cite{omnidepth} datasets are used for evaluation. Matterport \cite{matterport} dataset is not used since it requires additional data pre-processing which may provide different results according to how it is processed.
The following standard depth evaluation metrics are used to compare methods: absolute relative error (Abs.rel), squared relative error (Sq.rel), root mean square error (RMS), root mean square log error (RMSlog) and relative accuracy measures ($\delta$). Lower is better for Abs.rel, Sq.rel, RMS and RMSlog, whereas higher is better for $\delta$. Similar to \cite{midas,DPT}, we align the predicted depths using Eq.\ref{align} for all methods before measuring the errors.
For more details about the implementation, training, evaluation environment and additional experiments, refer to the the Technical, Code and Multimedia Appendix. 
\begin{table*}[t]
\renewcommand{\tabcolsep}{2mm}
\small
\centering
\caption{Quantitative comparison on Stanford dataset using the pre-trained baselines provided by each author. Numbers in \textbf{bold} indicate the best results.}
\begin{tabular}{cc||cccc|ccc}
\hline
Dataset & Method & AbsRel & Sq rel & RMS & RMSlog & $\delta \scalebox{0.9}{ < 1.25}$ & $\delta\scalebox{0.9}{ < 1.25}^2 $ & $\delta\scalebox{0.9}{ < 1.25}^3$ \\ \hline

\multirow{5}{*}{Stanford} & Omnidepth   & 0.1930 & 0.0042 & 0.0143 & 0.2691 &0.7663 & 0.9140 &  0.9635 \\
& Svsyn  & 0.1844 & 0.0039 & 0.0137 & 0.2596 & 0.7806 & 0.9220 & 0.9676 \\
& Bifuse & 0.1017 & 0.0019 & 0.0086 & 0.1783 & 0.9082 & 0.9722 & 0.9879 \\
& HoHoNet  & 0.0801 & 0.0016 & 0.0074 & 0.1577 & 0.9355 & 0.9803 & 0.9902 \\
& Ours w/ NLFB & \textbf{0.0666} & \textbf{0.0015} & \textbf{0.0066} & \textbf{0.1461} & \textbf{0.9531} & \textbf{0.9836} & \textbf{0.9910} \\ \hline 

\end{tabular}
\label{tab:Stanford}       
\end{table*}

\begin{table*}[t]
\renewcommand{\tabcolsep}{2mm}
\small
\centering
\caption{Further quantitative comparison using the baselines re-trained under the same training environment. Numbers in \textbf{bold} indicate the best results.}
\begin{tabular}{cc||cccc|ccc}
\hline
Dataset & Method & AbsRel & Sq rel & RMS & RMSlog & $\delta \scalebox{0.9}{ < 1.25}$ & $\delta\scalebox{0.9}{ < 1.25}^2 $ & $\delta\scalebox{0.9}{ < 1.25}^3$ \\ \hline

\multirow{3}{*}{Stanford3D} & Bifuse   & 0.0421 & 0.0160 & 0.2199 & 0.0842 & 0.9752 & 0.9948 &  0.9983 \\
& HoHoNet  & 0.0541 & 0.0237 & 0.2566 & 0.1030 & 0.9573 & 0.9915 & 0.9968 \\
& Ours w/ NLFB & \textbf{0.0344} & \textbf{0.0116} &\textbf{0.1921} & \textbf{0.0709} & \textbf{0.9843} & \textbf{0.9965} & \textbf{0.9987} \\ \hline \hline

\multirow{3}{*}{Matterport3D} & Bifuse & 0.0455 & 0.0186 & 0.2368 & 0.0859 & 0.9744 & 0.9943 & 0.9981 \\
& HoHoNet & 0.0612 & 0.0319 & 0.2950 & 0.1142 & 0.9523 & 0.9887 & 0.9960\\
& Ours w/ NLFB & \textbf{0.0364} & \textbf{0.0125} & \textbf{0.1963} & \textbf{0.0700} & \textbf{0.9852} & \textbf{0.9966} & \textbf{0.9988} \\ \hline \hline

\multirow{3}{*}{SunCG} & Bifuse & 0.0323 & 0.0141 & 0.2067 & 0.0739 & 0.9811 & 0.9954 & 0.9985 \\
& HoHoNet & 0.0518 & 0.0291 & 0.2789 & 0.1082 & 0.9587 & 0.9898 & 0.9967 \\
& Ours w/ NLFB & \textbf{0.0233} & \textbf{0.0076} & \textbf{0.1574} & \textbf{0.0534} & \textbf{0.9915} & \textbf{0.9979} & \textbf{0.9992} \\ \hline \hline

\multirow{3}{*}{Stanford} & Bifuse & 0.1237 & 0.0026 & 0.0114 & 0.2067 & 0.8684 & 0.9560 & 0.9823 \\
& HoHoNet & 0.1306 & 0.0028 & 0.0114 & 0.2138 & 0.8510 & 0.9511 & 0.9804\\
& Ours w/ NLFB & \textbf{0.0992} & \textbf{0.0018} & \textbf{0.0080} & \textbf{0.1717} & \textbf{0.9147} & \textbf{0.9764} & \textbf{0.9865} \\ 

\hline
\end{tabular}
\label{tab:Retrain}       
\end{table*}
\par{\noindent\textbf{Discussions on evaluation}}
Because training neural network is affected by numerous variables, it often becomes sensitive even to the small changes in hyper-parameters. Therefore, unifying the training environment of the all previous works may not result in the fair comparison. 
Actually, the training setup of each previous study differs significantly \cite{omnidepth,svsyn,bifuse,hohonet}, which makes it difficult to compare only the superiority of each method. For more fair and clear comparison, therefore, our methods are evaluated under the two settings. First, our method is compared with the pre-trained baselines provided by each author \cite{omnidepth,svsyn,bifuse,hohonet} in the open-source community at Section \ref{Comp_Pre2}. Because the performance of them is guaranteed by each author, this evaluation is fair and reproducible. Then, we compare our method with the baselines re-trained under the same training environment in Section \ref{Comp_SOTA}, which further clarifies the superiority of each method. 

\subsection{Comparison with the pre-trained baselines}
\label{Comp_Pre2}
In this subsection, the pre-trained baselines provided by each author are used for evaluation. Our model is trained using 3D60 \cite{omnidepth} and Stanford \cite{stanford} datasets, which are also used as training data in previous works.
\par{\noindent\textbf{Quantitative results}} Table \ref{tab:depth_results} shows the quantitative depth prediction results on 3D60 testset. Ours w/ FB represents the model using normal fusion blocks \cite{DPT}, while Ours w/ NLFB is the model using the proposed non-local fusion blocks. 
Except for the Abs.rel metric on Matterport3D, our method achieves significant improvements over previous works. We observe that pre-trained Bifuse produces biased results on some specific test splits, which results in worse quantitative results than others.
Meanwhile, Ours w/ NLFB provides better results than Ours w/ FB for all cases, which demonstrates the effectiveness of NLFB. Table \ref{tab:Stanford} shows the quantitative results on Stanford testset. For all metrics, Ours w/ NLFB provides the best results. 
\begin{table*}[!t]
\renewcommand{\tabcolsep}{2mm}
\small
\centering
\caption{Ablation study on Structure3D dataset. $\mathcal{L}_p$ is used for all methods. Numbers in \textbf{bold} indicate the best results.} 
\begin{tabular}{c|cc|c|c||cccc|ccc}
\hline

ID & $\mathcal{L}_I$ & $\mathcal{L}_D$ & $\mathcal{L}_{pix} + \mathcal{L}_{grad}$& NLFB &  AbsRel & Sq rel & RMS & RMSlog &  $\delta \scalebox{0.9}{ < 1.25}$ & $\delta\scalebox{0.9}{ < 1.25}^2 $ & $\delta\scalebox{0.9}{ < 1.25}^3$ \\ \hline
1 &\checkmark & & & & 0.1056 &  0.0120 & 0.0646 & 0.1611 & 0.9121 & 0.9792 & 0.9926  \\
2& \checkmark & \checkmark & & & 0.1048 & 0.0116 & 0.0631 & 0.1593 & 0.9151 & 0.9804 & 0.9928 \\
3& &  & \checkmark & & 0.0867 & 0.0086 & 0.0520 & 0.1403 & 0.9451 & 0.9855 &  0.9941 \\
4&\checkmark &  & \checkmark & & 0.0840 & 0.0079 & 0.0498 & 0.1360 & 0.9477 & 0.9870 & 0.9947 \\
5&\checkmark & \checkmark & \checkmark & & 0.0802 & 0.0071 & 0.0476 & 0.1302 & \textbf{0.9540} & \textbf{0.9887} & 0.9952 \\
6&\checkmark & \checkmark & \checkmark & \checkmark & \textbf{0.0781} & \textbf{0.0071} & \textbf{0.0470} & \textbf{0.1294} & 0.9537 & 0.9885 & \textbf{0.9953} \\
\hline
\end{tabular}

\label{tab:ablation_study}       
\end{table*}
\par{\noindent\textbf{Qualitative results}} Figure \ref{fig:depth_results} shows the qualitative depth estimation results. Omnidepth provides good results for small objects. Overall, however, unstable depth results are observed (\eg wall). HoHoNet produces stable depth results, but lacks detail. Small objects are not appropriately predicted, which is also reported as a weakness in their papers \cite{hohonet}. Meanwhile, it is observed that Bifuse provides fine qualitative results compared to other previous works. However, the results of Ours (/w NLFB) are much more accurate than Bifuse, which are even better than the ground truth for some cases. Holes and inaccurate depths are observed among the ground truth depths, whereas Ours provides stable and accurate depth results. 
\\

\subsection{Comparison with the re-trained baselines}
\label{Comp_SOTA}
In this subsection, we compare our method with the baselines re-trained under the same training environment to further clarify the superiority of our approaches.
 Bifuse and HoHoNet, which are the most recent studies and sharing similar training environment, are used for evaluation. 
 Following the training environment of Bifuse and HoHoNet, each method including ours is re-trained using 3D60 dataset with 512$\times$1024 resolutions. Table \ref{tab:Retrain} shows the quantitative results on the re-trained models. Compared to the results in Section \ref{Comp_Pre2}, some improvements are observed except at the Stanford testset, which is expected considering that they are trained with 3D60 dataset only with higher resolutions. As similar to the results in Section \ref{Comp_Pre2}, however, our approach provides the best results for all metrics at all testset. Considering the results of Tables \ref{tab:depth_results},\ref{tab:Stanford} and \ref{tab:Retrain} altogether, it is clear that our approach provides better results than others.
\begin{figure}[t]
\centering
\begin{minipage}{.24\textwidth}
\begin{subfigure}{\linewidth}
\includegraphics[width=.98\linewidth]{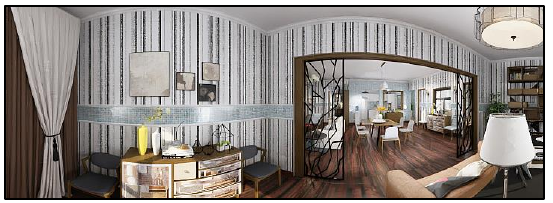}
\caption{Input}
\end{subfigure}
\end{minipage}%
\begin{minipage}{.24\textwidth}
\begin{subfigure}{\linewidth}
\includegraphics[width=.98\linewidth]{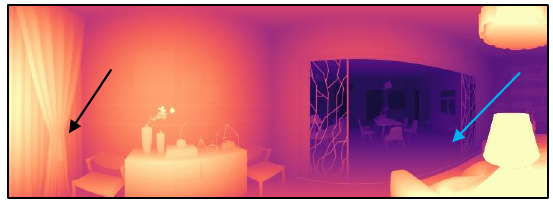}
\caption{ Ground truth }
\end{subfigure}
\end{minipage}%

\begin{minipage}{.24\textwidth}
\begin{subfigure}{\linewidth}
\includegraphics[width=.98\linewidth]{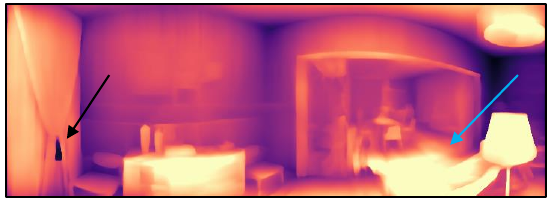}
\caption{ $\mathcal{L}_{pix} + \mathcal{L}_{grad}$ }
\end{subfigure}
\end{minipage}%
\begin{minipage}{.24\textwidth}
\begin{subfigure}{\linewidth}
\includegraphics[width=.98\linewidth]{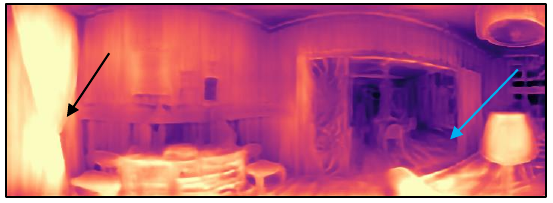}
\caption{$\mathcal{L}_{I} + \mathcal{L}_{D}$  }
\end{subfigure}
\end{minipage}%

\begin{minipage}{.24\textwidth}
\begin{subfigure}{\linewidth}
\includegraphics[width=.98\linewidth]{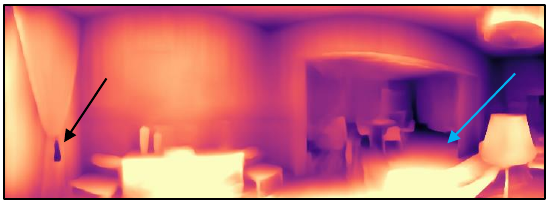}
\caption{$\mathcal{L}_{pix} + \mathcal{L}_{grad}+ \mathcal{L}_{I}$}
\end{subfigure}
\end{minipage}%
\begin{minipage}{.24\textwidth}
\begin{subfigure}{\linewidth}
\includegraphics[width=.98\linewidth]{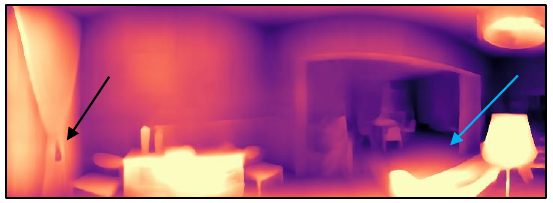}
\caption{$\mathcal{L}_{pix} + \mathcal{L}_{grad}+ \mathcal{L}_{I}+ \mathcal{L}_{D}$}
\end{subfigure}
\end{minipage}%
\caption{Studies on loss functions }
\label{fig:ablation_study}
\end{figure} 
\subsection{Ablation study}
\label{Ablation study}
In this section, we analyze the effect of each component of the proposed scheme through an ablation study on the Structure3D dataset \cite{structure3d}. Structure3D, a recently proposed dataset, is not used in the training procedure. Therefore, it is suitable to demonstrate our arguments on the proposed loss functions (\eg perform well at unexposed data). Table \ref{tab:ablation_study} shows the quantitative results when proposed schemes are added gradually. $\mathcal{L}_p$ is used in all cases, though it is omitted in Table \ref{tab:ablation_study} for better visualization. When only self-supervised losses are applied (ID 1 and 2), the results are not good as expected. However, when self-supervised losses are used with supervised losses (ID 4 and 5), the performance increases dramatically compared to the cases when only supervised losses are applied (ID 3). In this case, $\mathcal{L}_D$ plays an important role when both losses are used together (ID 4 and 5). This results show that self-supervised learning actually improves the depth estimation results when combined with supervised learning. Also, it is observed that NLFB improves the performance further (ID 6).

Figure \ref{fig:ablation_study} shows the qualitative result of the schemes in Table \ref{tab:ablation_study}. The model trained only with supervised losses in Figure \ref{fig:ablation_study} (c) produces unsatisfactory results for some areas. The black arrows indicate where the model predicts a decoration in front of the curtain as a defect. Therefore, these parts are predicted as holes, which are often found in ground truth depths. This indicates that the model is highly affected by inaccurate ground truth depths. The blue arrows indicate cases where the model fails to distinguish a sofa from the floor, which occurs because the model was not exposed to such cases during the training procedure. On the other hand, the model trained with self-supervised losses in Figure \ref{fig:ablation_study} (d) was able to recognize the black object at the curtain appropriately, and distinguish the sofa from the floor. However, it produces unstable depths overall. When $\mathcal{L}_I$ is used with supervised losses as shown in Figure \ref{fig:ablation_study}(e), the problems in Figure \ref{fig:ablation_study} (c) are mitigated according to the results shown in  Figure \ref{fig:ablation_study} (d). When $\mathcal{L}_D$ is applied in addition, as shown in Figure \ref{fig:ablation_study} (f), the model was able to distinguish the objects properly. 
\begin{figure}[!t]
\centering
\begin{minipage}{.24\textwidth}
\begin{subfigure}{\linewidth}
\includegraphics[width=.98\linewidth]{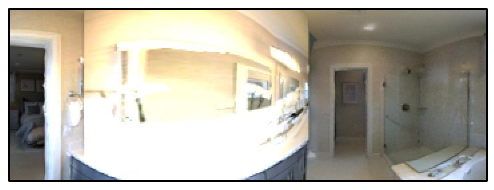}
\caption{Input}
\end{subfigure}
\end{minipage}%
\begin{minipage}{.24\textwidth}
\begin{subfigure}{\linewidth}
\includegraphics[width=.98\linewidth]{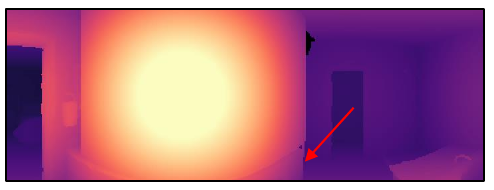}
\caption{ Ground truth }
\end{subfigure}
\end{minipage}%

\begin{minipage}{.24\textwidth}
\begin{subfigure}{\linewidth}
\includegraphics[width=.98\linewidth]{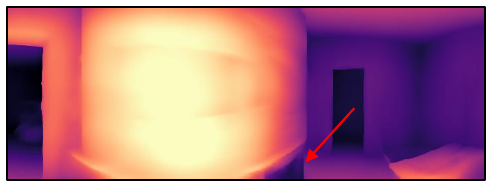}
\caption{ Fusion block }
\end{subfigure}
\end{minipage}%
\begin{minipage}{.24\textwidth}
\begin{subfigure}{\linewidth}
\includegraphics[width=.98\linewidth]{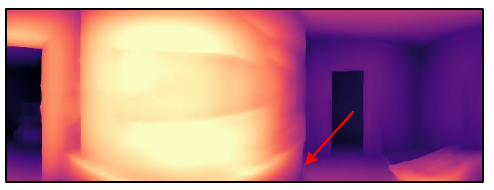}
\caption{Non-local fusion block}
\end{subfigure}
\end{minipage}%
\caption{Effect of non-local fusion block}
\label{fig:ablation_nlfb}
\end{figure} 
Figure \ref{fig:ablation_nlfb} shows the effect of a non-local fusion block in more detail. The fusion block fails to recognize the wall as a single object, and therefore, reconstructs undesirable depths (red arrows). On the other hand, the non-local fusion block reconstructs the depth well. This demonstrates that NLFB makes the network to continue to see the features with a wider view when reconstructing the depths.

\section{Conclusion}
In this paper, we introduce a self-supervised learning scheme, a joint objective function, and a non-local fusion block, in an effort to address the problems found in studies of EI depth estimations. Through the proposed scheme, significant improvements over prior works are achieved, and the benefits of each proposed method are also analyzed. We believe that each contribution not only affects the EI depth estimation research but also provides insight for those involved in studies of other vision tasks.

\section*{Acknowledgment}
This work was supported by Samsung Research Funding Center of Samsung Electronics under Project Number SRFC-IT1702-54.

{\small
\bibstyle{aaai22}
\bibliography{egbib}
}

\clearpage
\renewcommand{\thefigure}{\Alph{figure}}
\setcounter{figure}{0}
\renewcommand{\thetable}{\Alph{table}}
\setcounter{table}{0}
\renewcommand{\thesection}{\Alph{section}}
\setcounter{section}{0}
\noindent\textbf{\LARGE{Technical Appendix}}
\section{Detailed experimental environment}
\subsection{Network architecture}
\par
\begin{figure}[h]
\centering

\begin{minipage}{.49\textwidth}
\begin{subfigure}{\linewidth}
\includegraphics[width=.98\linewidth]{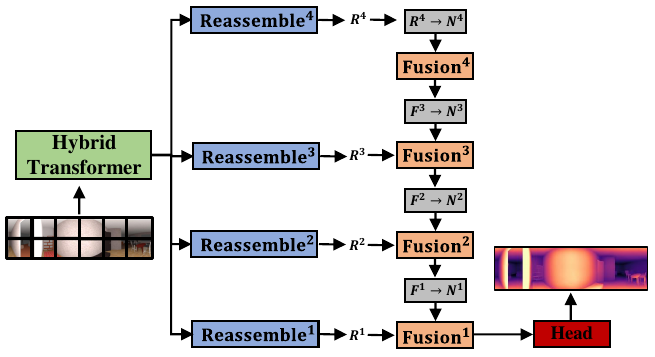}
\caption{Non-local DPT}
\end{subfigure}
\end{minipage}%

\begin{minipage}{.45\textwidth}
\begin{subfigure}{\linewidth}
\includegraphics[width=.98\linewidth]{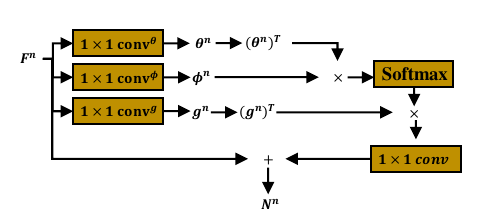}
\caption{Non-local block}
\end{subfigure}
\end{minipage}%

\caption{Network architecture}
\label{fig:supp_network}
\end{figure}

{\noindent\textbf{Depth estimation network}}
 The overall architecture of non-local DPT is illustrated in Figure \ref{fig:supp_network} (a). We use the ViT-hybrid \cite{vision_transformer} model as an architecture of Hybrid Transformer, and the DPT-hybrid \cite{DPT} model as an architecture of reassemble, fusion and head blocks. The non-local fusion block is made by concatenating non-local block, as illustrated in Figure \ref{fig:supp_network} (b), with fusion block. 

\par{\noindent\textbf{Pose estimation network}}
We use PoseCNN \cite{monodepth2} as our PoseNet architecture . 

\subsection{Training dataset}
\par{\noindent\textbf{Supervised learning}}
For supervised learning flows, we use 3D60 dataset \cite{omnidepth} (Matterport3D, Stanford3D and SunCG) and Stanford dataset \cite{stanford}, which are also used in previous works \cite{omnidepth,svsyn,bifuse,hohonet}.
\par{\noindent\textbf{Self-supervised learning}}
For self-supervised learning flows, we captured videos using a single \ang{360} camera. We fix the camera on the horizontal plane of a moving device (\eg dolly), which is a common means of capturing EI videos. Then, we capture the videos by only moving the device straight forward for simplicity. By fixing a camera onto a horizontal flat plane, the video is gravity aligned \cite{Slicenet}. By doing this, we could use videos directly as training data without calibration. Because all that is necessary here is capturing the videos, we could acquire 40,000 video sequences from 4 different locations with little effort, which can also be easily obtained by others. 

\begin{table*}[t]
\renewcommand{\tabcolsep}{2mm}
\small
\centering
\caption{Quantitative comparison between self-supervised learning studies. Numbers in \textbf{bold} indicate the best results.}
\begin{tabular}{ccc||cccc|ccc}
\hline
Testset & Method & Training data type & AbsRel & Sq rel & RMS & RMSlog & $\delta < 1.2$ & $\delta < 1.2^2 $ & $\delta < 1.2^3$ \\ \hline

\multirow{3}{*}{Structure3D} & EBS  & Stereo image & 0.1225 & 0.0154 & 0.0727 & 0.1801 & 0.8983 & 0.9723 &  0.9894 \\
& SvSyn & Stereo image& 0.1142 & 0.0134 & 0.0679 & 0.1713 & 0.9058 & 0.9749 &  0.9908 \\
& Ours & Video & \textbf{0.1048} & \textbf{0.0116} & \textbf{0.0631} & \textbf{0.1593} & \textbf{0.9151} & \textbf{0.9804} & \textbf{0.9928} \\ \hline 
\end{tabular}
\label{tab:SS_results}       
\end{table*}
\begin{figure*}[t]
\centering

\begin{minipage}{.333\textwidth}
\begin{subfigure}{\linewidth}
\includegraphics[width=.98\linewidth]{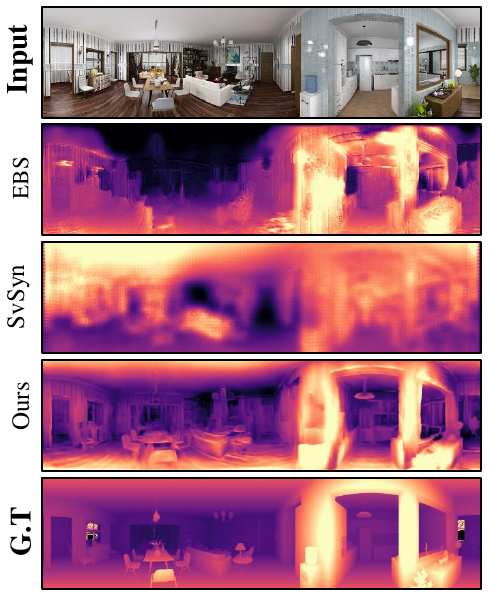}
\end{subfigure}
\end{minipage}%
\begin{minipage}{.333\textwidth}
\begin{subfigure}{\linewidth}
\includegraphics[width=.98\linewidth]{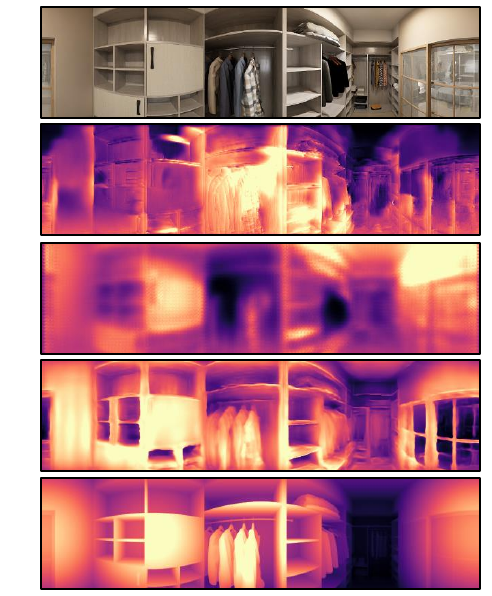}
\end{subfigure}
\end{minipage}%
\begin{minipage}{.333\textwidth}
\begin{subfigure}{\linewidth}
\includegraphics[width=.98\linewidth]{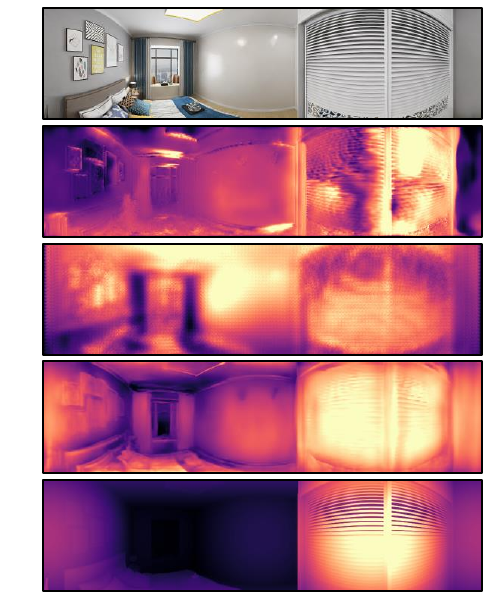}
\end{subfigure}
\end{minipage}%
\caption{Qualitative comparison between self-supervised learning studies }
\label{fig:supp_self_results}
\end{figure*} 
\subsection{Training details}
\par{\noindent\textbf{Data pre-processing}}
Because \ang{360} camera covers scenes at all angles, equipment (\eg tripods) and/or other sensors (\eg depth sensor) are inevitably captured as well when capturing the images/videos \cite{Lowcost, omnidepth}. These unexpected artefacts are not defined using depth and camera motions because they are fixed in the scenes regardless of the camera movements. This causes some confusing in the network when it is trained in a self-supervised manner. Therefore, floor and ceiling parts (a \ang{45} $fov_y$ region for each top/bottom area) of the video data are cropped before training phase (experiments of Tables \ref{tab:depth_results} and \ref{tab:ablation_study}), or not used when calculating the loss during training phase (experiments of Tables \ref{tab:Stanford} and \ref{tab:Retrain}). 
\par{\noindent\textbf{Training environment}}
As stated in the main paper, we initialize the weights of the non-local DPT using a model trained with rectilinear depths excluding the non-local blocks. Then, we train the network for ten epochs using only self-supervised learning flows to pre-train the pose network. Because our video data is captured by only moving the camera straight forward as stated above, we were able to further simplify the camera motions into a single variable $\delta x$, which makes training much easier. Subsequently, we re-initialize the weights of the non-local DPT using the model trained with rectilinear depths, after which, we train the network using both the supervised and self-supervised learning flows for ten epochs with a fixed pose network. Because the process of back-propagating two flows in a single training step is not efficient due to GPU memory limitations, we randomly execute one of the flows during each training step. We adjust the scales of ground truth and the estimated depth of supervised learning flows to be in the range of $(0,1)$. The ground truth is normalized to $(0,1)$, whereas the estimated depths are clamped to have value between $(0,1)$ at the head block in Figure \ref{fig:supp_network}.
 For experiments in Tables \ref{tab:depth_results} and \ref{tab:ablation_study}, we use top/bottom cropped $128\times 512$ 3D60 and video data. 
 For experiments in Table \ref{tab:Stanford}, we use $512\times 1024$ Stanford data \cite{stanford} and  $256\times 512$ video data. For experiments in Table \ref{tab:Retrain}, we use resized $512\times 1024$ 3D60 data and $256\times 512$ video data.
For all experiments, Adam optimizer with a learning rate of 0.0002 and (0.5,0.999) beta values is used. For more details, readers can refer to the code appendix.

\subsection{Evaluation details}

Following the evaluation environment of Bifuse \cite{bifuse} and HoHoNet \cite{hohonet}, we resize the input image to $512 \times 1024$ for evaluation. 
Because the top and bottom parts of the EIs contain similar depths for most scenes, as stated above, evaluating the depth for the middle part reflects the actual performance better. Therefore, we do not measure the errors of the top/bottom parts of the estimated depths (a \ang{45} $fov_y$ region for each top/bottom area) when evaluating each methods. The scale and shift of each estimated depth of each method is aligned with Eq.\ref{align} before measuring the errors for a fair comparison.

\section{Additional experiments}

\subsection{Comparison with self-supervised learning studies}
To demonstrate the superiority of our proposed self-supervised learning scheme, we compare our method with the other self-supervised learning studies. Among them, EBS \cite{EBS} and SvSyn \cite{svsyn} are compared. Because 360SD-Net \cite{360sdnet} requires stereo top-down image pairs for test and \cite{360self} needs the conversion to cubemap projection, they are not compared. Meanwhile, EBS uses KITTI \cite{kitti} dataset, SvSyn uses Matterport3D and Stanford3D dataset, and we use videos for training data. Therefore, comparison on common dataset (Matterport3D, Stanford3D and SunCG) is not fair because SvSyn is exposed during the training procedure. Therefore, we compare the results on Structure3D \cite{structure3d} dataset only which is not used by any other methods. Table \ref{tab:SS_results} shows the quantitative results of each method. Although our method is trained using videos which is far more challenging, we achieve the best results among the self-supervised learning studies. Figure \ref{fig:supp_self_results} shows the qualitative results of each method. EBS and SvSyn provide poor results, whereas our approach shows much more accurate detphs.

\subsection{Analysis on training setup}
In this subsection, the empirical analysis on our training setup, which are described in Section \ref{sec:NLDPT}, are explained. Table \ref{tab:training_setup} shows the quantitative results on the Structure3D test set when different training setup is applied. We observe that learning from the scratch using EI depth data fails to learn appropriate features as shown in the first row of Table \ref{tab:training_setup}. The network was not able to learn any useful features when a pre-trained model on large-scale dataset is not provided. This is expected because numerous works have already demonstrated the necessity of fine-tuning for the transformer. Meanwhile, we also observe that robust adjustment on ground truth and the estimated depths also affect the training significantly. Without robust adjustment, scale mismatches between supervised and self-supervised learning flows dominate the total loss, which keeps the network from being trained successfully, as shown in the second row of Table \ref{tab:training_setup}. When full training set up is applied, the training is done successfully as shown in the third row of Table \ref{tab:training_setup}.

\begin{table}[h]
\renewcommand{\tabcolsep}{2mm}
\small
\centering
\caption{Effect of training setup.}
\begin{tabular}{c||ccc|c}
\hline
Training setup & AbsRel & Sq rel & RMS &  $\delta < 1.25$  \\ \hline

w/o fine-tuning & 1.000 & 0.4276 & 0.4755 & 0.0000 \\
w/o robust adjustment & 0.3793 & 0.0985 & 0.2135 & 0.4199\\
Baseline & 0.0781 & 0.071 & 0.0470 & 0.9537 \\ \hline 

\end{tabular}
\label{tab:training_setup}       
\end{table}

\begin{figure}[h]
\centering
\begin{minipage}{.48\textwidth}
\begin{subfigure}{\linewidth}
\includegraphics[width=\linewidth]{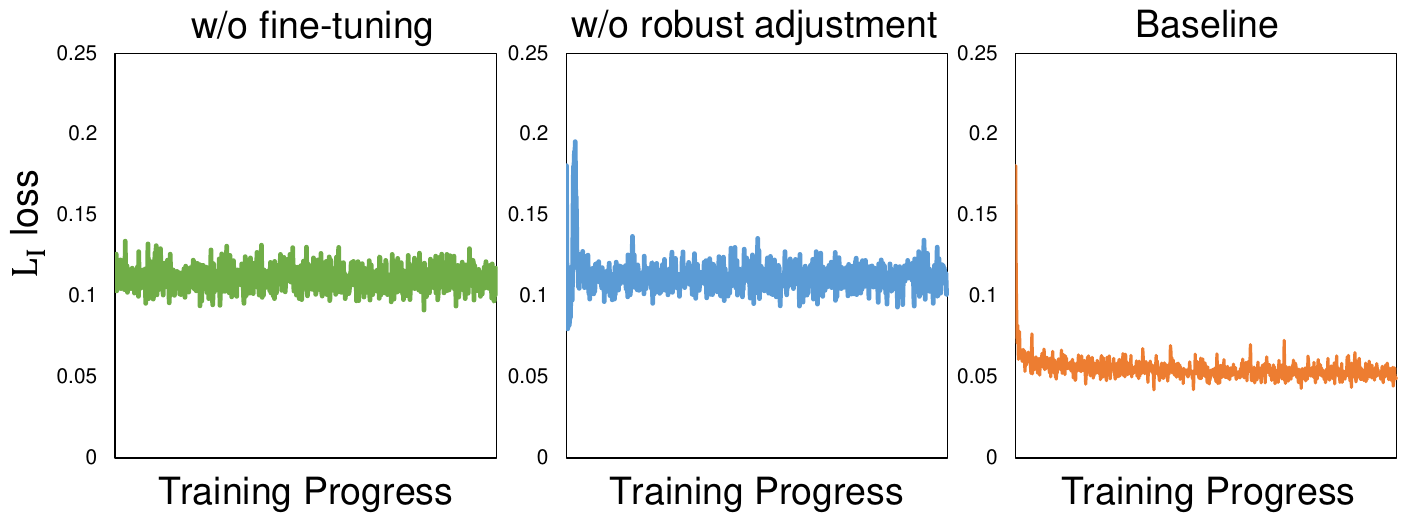}
\end{subfigure}
\end{minipage}%
\caption{Evolution of $\mathcal{L}_I$ loss over training }
\label{fig:supp_graph}
\end{figure}

Figure \ref{fig:supp_graph} shows the evolution of $\mathcal{L}_I$ loss over training (10 epochs) for each training setup. Baseline, which uses full training setup, converges properly. When training is done without robust adjustment, loss oscillates harshly in the early phase of training and does not converge to the appropriate point. Similar tendency is also observed when training is done from the scratch. 

\subsection{Failure cases of self-supervised learning}
\begin{figure}[h]
\centering
\begin{minipage}{.45\textwidth}
\begin{subfigure}{\linewidth}
\includegraphics[width=\linewidth]{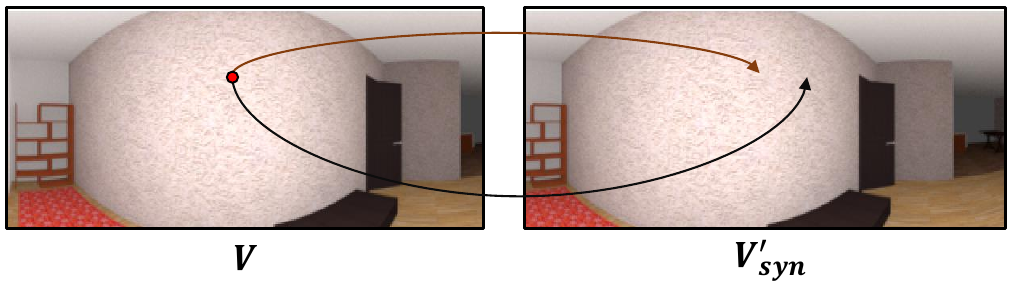}
\end{subfigure}
\end{minipage}%
\caption{An example of non-unique solution in self-supervised learning}
\label{fig:self_limit}
\end{figure}

As mentioned in the main paper, self-supervised learning on depths often delivers wrong or non-unique solutions. Figure \ref{fig:self_limit} shows one of the cases that has non-unique solutions. Consider the situation where $V'_{syn}$ is synthesized from $V$ using two different depth values. 
The pixel of red dot in $V$ is synthesized differently according to the two different depth values, which are denoted by brown and black arrows. As shown in Figure \ref{fig:self_limit}, however, difference in depth values of red dot provides similar synthesized results because the wall has uniform textures. 
 In this case, different depth values may receive similar feedbacks from image consistency loss, which results in non-unique solutions. If so, depth can be multi-valued and this causes unstable results.

\begin{figure}[h]
\centering
\begin{minipage}{.45\textwidth}
\begin{subfigure}{\linewidth}
\includegraphics[width=\linewidth]{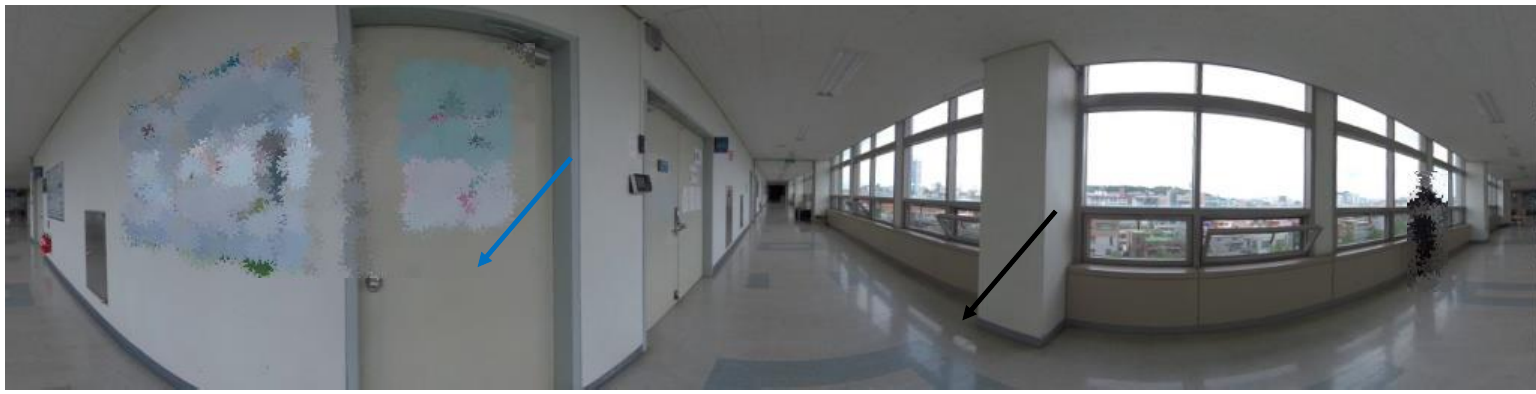}
\end{subfigure}
\end{minipage}%

\begin{minipage}{.45\textwidth}
\begin{subfigure}{\linewidth}
\includegraphics[width=\linewidth]{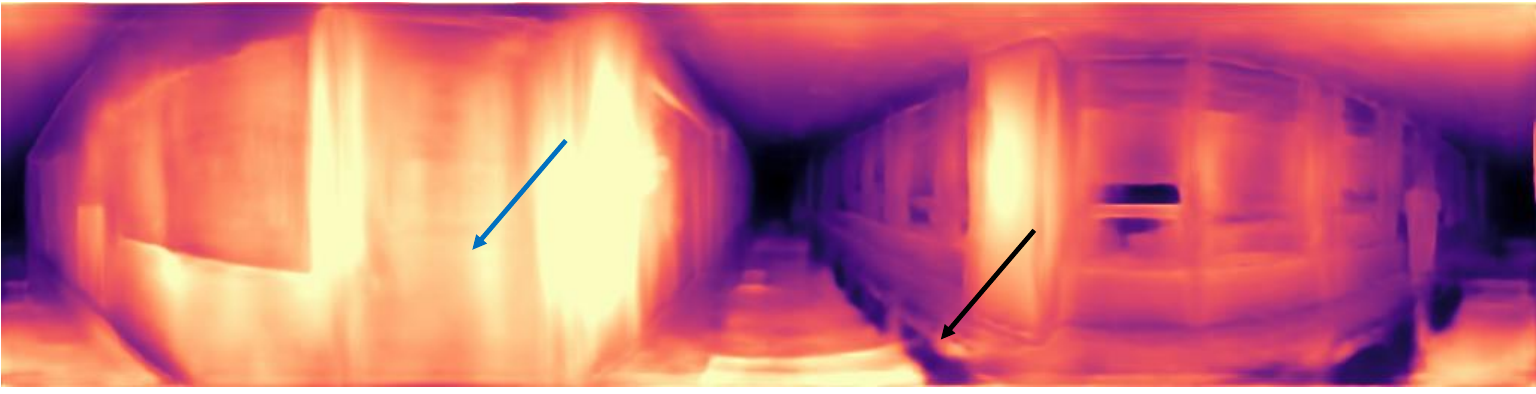}
\end{subfigure}
\end{minipage}%
\caption{Failure cases of self-supervised learning}
\label{fig:fail_case}
\end{figure}

 Figure \ref{fig:fail_case} shows the failure cases of self-supervised learning. The blue arrows indicate where the model produce the unstable depth due to uniform texture of the wall and door. The black arrows indicate where the model fails to predict the depth of light reflected area. Although those failures can be alleviated via supervised learning as we argued in our paper, they should be resolved fundamentally for more accurate depth estimation, which we left to the future.

\subsection{Additional qualitative results}
\begin{figure*}[!t]
\centering
\begin{minipage}{.333\textwidth}
\begin{subfigure}{\linewidth}
\includegraphics[width=.98\linewidth]{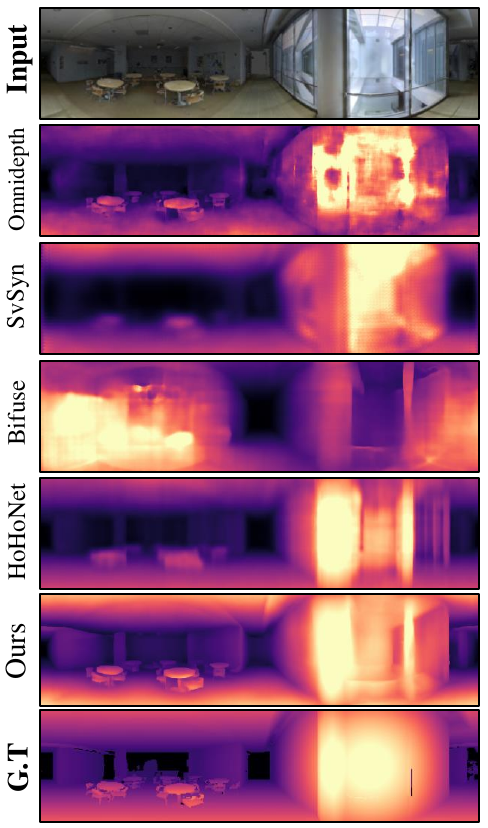}
\end{subfigure}
\end{minipage}%
\begin{minipage}{.333\textwidth}
\begin{subfigure}{\linewidth}
\includegraphics[width=.98\linewidth]{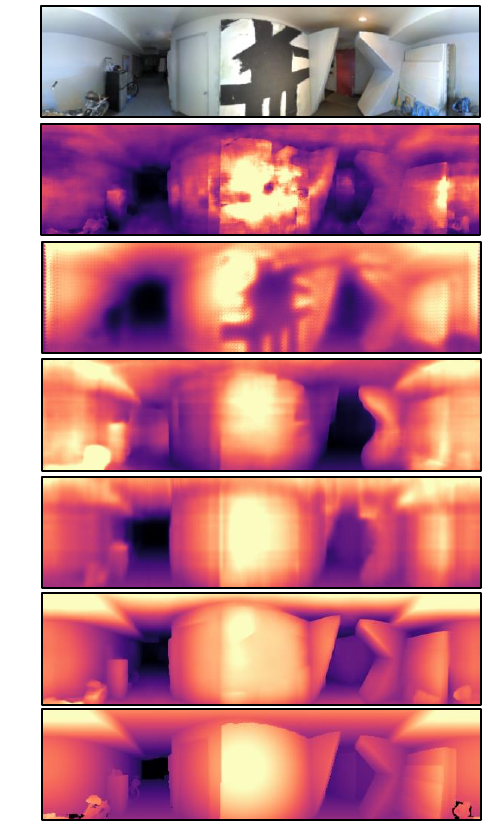}
\end{subfigure}
\end{minipage}%
\begin{minipage}{.333\textwidth}
\begin{subfigure}{\linewidth}
\includegraphics[width=.98\linewidth]{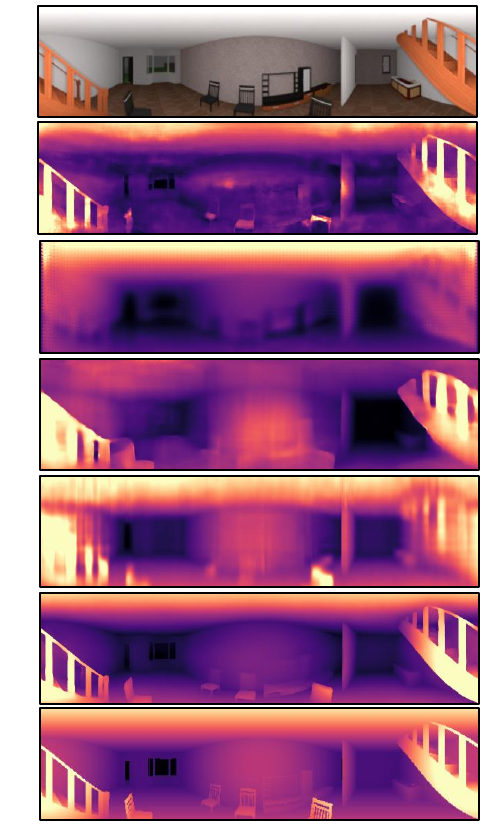}
\end{subfigure}
\end{minipage}%

\begin{minipage}{.333\textwidth}
\begin{subfigure}{\linewidth}
\includegraphics[width=.98\linewidth]{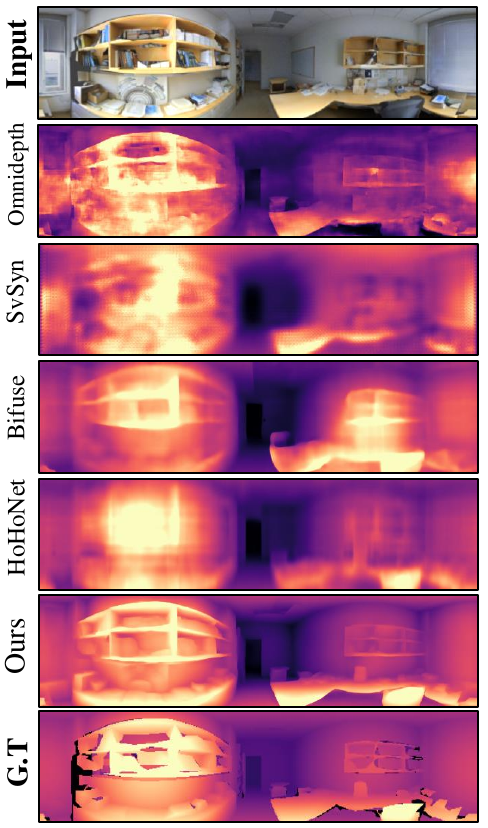}
\caption{Stanford3D}
\end{subfigure}
\end{minipage}%
\begin{minipage}{.333\textwidth}
\begin{subfigure}{\linewidth}
\includegraphics[width=.98\linewidth]{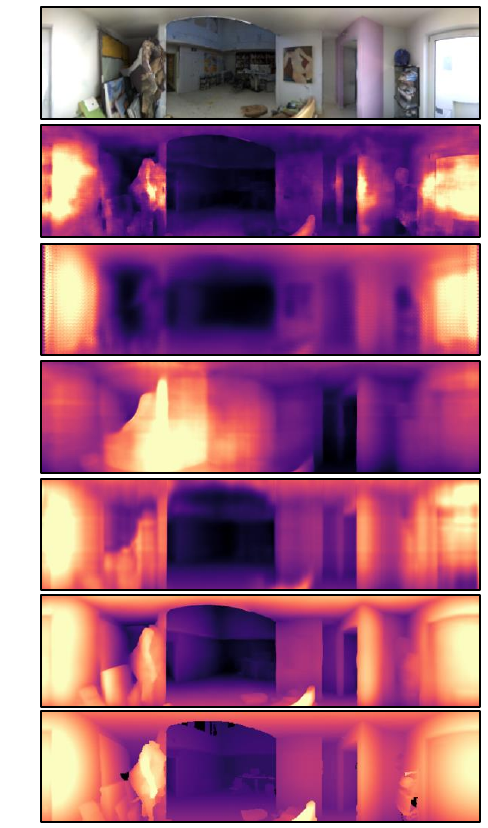}
\caption{Matterport3D}
\end{subfigure}
\end{minipage}%
\begin{minipage}{.333\textwidth}
\begin{subfigure}{\linewidth}
\includegraphics[width=.98\linewidth]{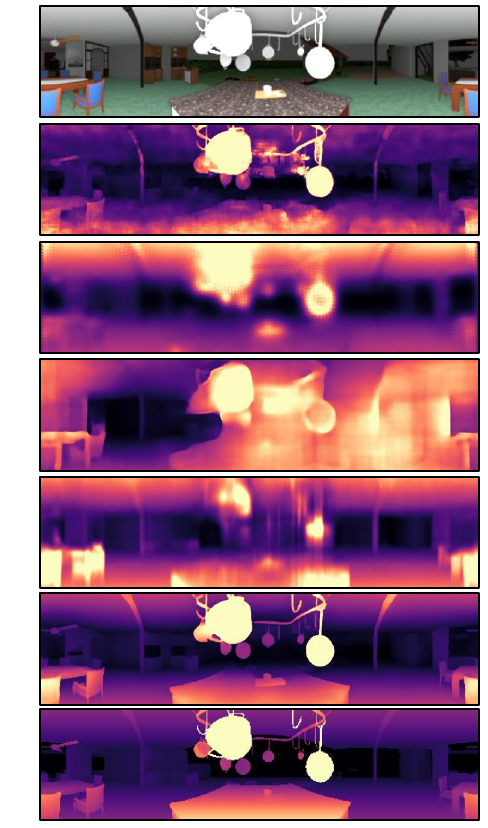}
\caption{SunCG}
\end{subfigure}
\end{minipage}%

\caption{Additional qualitative results on Stanford3D, Matterport3D and SunCG testset.  }
\label{fig:supp_depth_results}
\end{figure*}

Figure \ref{fig:supp_depth_results} shows the additional qualitative results on Stanford3D, Matterport3D and SunCG testsets for each pre-trained baselines.

\end{document}